\documentclass[twocolumn]{article}

\usepackage[ruled,vlined,linesnumbered,noend]{algorithm2e}
\usepackage{amsmath}
\usepackage{amssymb}
\DeclareMathOperator*{\argmax}{arg\,max}
\usepackage{enumitem}
\usepackage{soul}
\usepackage{hyperref}
\usepackage[margin=2cm]{geometry}
\usepackage{graphicx}
\usepackage[all]{nowidow}

\begin{document}

\title{A Genetic Fuzzy System for Interpretable and Parsimonious Reinforcement Learning Policies\,\footnote{© ACM 2021. This is the author's version of the work. It is posted here for your personal use. Not for redistribution. The definitive Version of Record was published in \textit{GECCO '21: Proceedings of the Genetic and Evolutionary Computation Conference Companion}, \url{http://dx.doi.org/10.1145/3449726.3463198}.}}

\author{Jordan T. Bishop\\
        The University of Queensland, Australia\\
        \texttt{j.bishop@uq.edu.au}
        \and
        Marcus Gallagher\\
        The University of Queensland, Australia\\
        \texttt{marcusg@uq.edu.au}
        \and
        Will N. Browne\\
        Queensland University of Technology, Australia\\
        \texttt{will.browne@qut.edu.au}}

\date{}
\maketitle

\section*{Abstract}
Reinforcement learning (RL) is experiencing a resurgence in research interest, where Learning Classifier Systems (LCSs) have been applied for many years. However, traditional Michigan approaches tend to evolve large rule bases that are difficult to interpret or scale to domains beyond standard mazes. A Pittsburgh Genetic Fuzzy System (dubbed Fuzzy MoCoCo) is proposed that utilises both multiobjective and cooperative coevolutionary mechanisms to evolve fuzzy rule-based policies for RL environments. Multiobjectivity in the system is concerned with policy performance vs. complexity. The continuous state RL environment Mountain Car is used as a testing bed for the proposed system. Results show the system is able to effectively explore the trade-off between policy performance and complexity, and learn interpretable, high-performing policies that use as few rules as possible.

\section{Introduction}
Genetics-Based Machine Learning (GBML) \cite{kovacs_genetics-based_2012} has a long history of being applied to reinforcement learning (RL) problems where new methods are needed to take advantage of the renewed interest in such domains.
Genetic Fuzzy Systems (GFSs) \cite{cordon_genetic_2001} are a type of GBML that aim to evolve fuzzy rule-based systems (FRBSs).
Also under the umbrella of GBML, Learning Classifier Systems (LCSs) are a family of evolutionary rule-based systems that create solutions to machine learning problems.

Within both families, there are two broad types of systems that represent different ways to solve a problem: the Michigan and Pittsburgh approaches. Both approaches utilise population-based evolutionary mechanisms. In a Michigan system, each individual in the population is an element of the solution; all individuals act in ensemble to create the entire solution. In contrast, a Pittsburgh system treats each individual in the population as an entire solution to the problem \cite{cordon_genetic_2001, urbanowicz_introduction_2017}. Within GFSs, a wide array of works have focused on the Pittsburgh approach \cite{cordon_genetic_2001, herrera_genetic_2008}, while in the LCS literature the predominant paradigm is Michigan \cite{urbanowicz_introduction_2017}.

Both LCSs and GFSs can be applied to RL problems. LCSs were originally designed to perform RL, and much work has been done in this area already, particularly in maze-like environments, e.g. \cite{lanzi_xcs_2005, loiacono_recursive_2008}. In contrast, most GFS work has focused on supervised learning: classification or regression \cite{herrera_genetic_2008}, with some work being done on ``control'' problems, e.g. \cite{homaifar_simultaneous_1995}. However, such control problems are often not formulated under the RL framework; this framework prescribes problems that are multi-step, involve delayed rewards, and are characterised by two fundamental issues: the explore-exploit dilemma and temporal credit assignment. Michigan and Pittsburgh systems address these issues at different levels of abstraction. Michigan systems learn in an online fashion, and they address both issues at the level of individual state-action pairs within a stream of experience. Particularly in problems where exploration is difficult and/or reward signals are sparse, this can be difficult to achieve. On the other hand, Pittsburgh systems assign credit to entire solutions, and address the explore-exploit problem in the more abstract policy parameter space.

Generally, there is a lack of work applying LCSs to common environments from the RL literature that are not maze-like, e.g. Mountain Car or Cart Pole \cite{sutton_reinforcement_2018}, an exception being \cite{stein_xcs_2020}. Such environments often have continuous state spaces. Since LCSs prescribe a paradigm of learning rather than a specific algorithm, they enable the representation of rule conditions to be flexibly chosen to suit the problem domain. For continuous domains, there are a variety of choices available, some examples being hyperrectangles \cite{loiacono_recursive_2008}, hyperellipsoids \cite{butz_kernel-based_2005}, and fuzzy logic \cite{valenzuela-rendon_reinforcement_1998}. Fuzzy logic attempts to perform inference in a way that better emulates how a human expert may solve a problem by including degrees of truth rather than absolute values. It is an attractive representation to use if the purpose of the system is to produce a human-understandable explanation of how a problem is solved.

An issue for both LCSs and GFSs is how to deal with the complexity of rule bases that are evolved in order to prefer parsimonious (low complexity) models. In the Michigan approach, a post hoc compaction mechanism is often employed to remove rules that do not contribute much to the solution \cite{urbanowicz_introduction_2017}. In contrast, more options are available for Pittsburgh systems, some being: i) limiting the size of candidate solutions, ii) employing fitness penalisation based on complexity (\cite{bacardit_bloat_2007} is an approach that uses the Minimum Description Length principle in this manner), iii) multiobjective (MO) formulation of solution performance vs. complexity \cite{ishibuchi_multiobjective_2007,llora_bounding_2003}. The third strategy is particularly attractive if the practitioner desires to understand the trade-off between performance and complexity; and to understand \textit{how many rules are needed to achieve a given performance value}.

Therefore, the first objective of this work is to address an RL problem that incorporates both i) difficult exploration, and ii) a continuous state space (a candidate is Mountain Car, see justification Section~\ref{sec:mc}). Objective two is to understand the trade-off between rule base complexity and performance, through employing a Pittsburgh GFS that performs MO optimisation of FRBSs. Finally, since an FRBS can be naturally decomposed into a rule base (RB) and a data base (DB) (see Section~\ref{sec:frbs_structure}), the third objective is to show that cooperative coevolution (CoCo) can be employed to jointly optimise the RB and DB. Thus, the overall aim of this work is to develop a Pittsburgh GFS that utilises CoCo and MO mechanisms to produce parsimonious and interpretable policies for RL problems. As a proof of concept, we show that this system is able to produce compact and interpretable policies for the Mountain Car problem.
\section{Background}
\subsection{Reinforcement Learning}
In RL, an agent interacts with an (episodic) environment $\mathcal{E}$ to maximise the expected amount of cumulative reward it receives. Let $\mathcal{E}= (S, A, P, R, \gamma, t_\text{max})$, where $S$ is the state space, $A$ is the action space, $P(s'|s,a)$ is the transition function, $R(s,a,s')$ is the reward function, $0 \le \gamma \le 1$ is the discount factor, and $t_\text{max}$ is the maximum number of episode time steps \cite{sutton_reinforcement_2018}. In this work, we assume $S$ is continuous and $A$ is discrete.

The agent takes the form of a policy, which is a mapping $\pi: S \rightarrow A$. A common way to address RL problems is for the agent to construct an \textit{action-value function} $Q: S \times A \rightarrow \mathbb{R}$ which represents the expected cumulative reward obtainable from each state-action pair. A policy can then be constructed by acting greedily with respect to $Q$. This is the approach followed by Michigan systems.

An alternative way to address the problem is to treat the environment as a black box and perform \textit{direct policy search}; an approach taken by Pittsburgh systems. In this view, the agent receives feedback about its performance via a collective sum of discounted rewards, termed the \textit{return}: $G$ \cite{sutton_reinforcement_2018}. The task is to construct a policy that directly maximises the \textit{expected return}, without decomposing the return into individual rewards. The expected return (performance) of a policy is measured over a set of initial states $Z$ of cardinality $\eta$ drawn from an initial state space $S_I \subseteq S$. Using this formulation, the performance of a policy, abbreviated $\text{perf}$, can be measured as:
\begin{equation}
    \text{perf} = \frac{1}{\eta}\sum_{z \in Z}^{} G(z)
    \label{eqn:perfEval}
\end{equation}
where $G(z)$ is the return yielded by performing a rollout of the policy, starting at initial state $z$.

\subsection{Fuzzy Rule-Based Systems}
We use the following terminology when discussing aspects of fuzzy reasoning:
\textit{Linguistic variable} --- analogous to an environmental feature; includes both the name of the feature and a fuzzy partition along its domain.
A fuzzy partition is composed of multiple fuzzy sets.
\textit{Fuzzy set} --- defined by a membership function along the domain of a linguistic variable; has an associated name or \textit{linguistic value} to linguistically describe the set \cite{cordon_genetic_2001}.

\subsubsection{FRBS Structure}
\label{sec:frbs_structure}
The specific FRBS type considered is a zero-order Takagi-Sugeno-Kang system \cite{takagi_fuzzy_1985}. The FRBS is composed of two components, that together form the knowledge base: the rule base (RB) and data base (DB) \cite{cordon_genetic_2001}. The rule base contains the fuzzy rules that act in the context of the fuzzy partitions contained in the DB.
In the RB, we use fuzzy rules that are individually expressed in Conjunctive Normal Form (CNF) \cite{cordon_genetic_2001}. Assuming the dimensionality of $S$ is $d$ and there are $k$ possible actions ($\vert A \vert = k$), each rule has the structure:
\begin{center}
$\text{IF}\ x_1\ \text{is}\ \widetilde{L_1}=\{L_{(1,1)}\ \text{or}\ \ldots\ \text{or}\ L_{(1,m_1)}\}\ \text{and}\ \ldots$ \\
$\text{and}\ x_d\ \text{is}\ \widetilde{L_d}=\{L_{(d,1)}\ \text{or}\ \ldots\ \text{or}\ L_{(d,m_d)}\}$ \\
$\text{THEN}\ y_1 = \alpha_1, \ldots, y_k = \alpha_k$
\end{center}
where, for $i \in \{1, \ldots, d\}$:
\begin{itemize}
    \item $x_1, \ldots, x_d$ --- components of input vector $x$; linguistic variables
    \item $m_i$ --- num. linguistic values belonging to $i^{th}$ linguistic variable
    \item $L_{(i,j)}, j \in \{1, \ldots, m_i\}$ --- $j^{th}$ linguistic value of $i^{th}$ linguistic variable
    \item $\widetilde{L_i}$ --- non-empty set of linguistic values for $i^{th}$ linguistic variable
    \item $y_1, \ldots, y_k$ --- components of consequent vector $y$
    \item $\alpha_1, \ldots, \alpha_k$ --- voting weights for each action in consequent
\end{itemize}
$\alpha_1, \ldots, \alpha_k$ are constrained to be either 0 (inactive) or 1 (active), with exactly one weight active in every rule, all others inactive, i.e. a one-hot encoding. Such a scheme represents each rule voting (fully) for a single action in its consequent. When writing rule consequents we simply specify the action whose weight is active ($a\ is\ k$).
This type of rule allows for flexible levels of generalisation. Selecting all linguistic values of a linguistic variable is equivalent to a ``don't care'', denoted by \#. Note that it is not possible to select \textit{zero} linguistic values; as stated above the set of linguistic values must be non-empty. As an example, assuming that $d=2,\ k=2,\ m_1=m_2=3$, the following CNF rule generalises partially over the first feature and fully over the second feature:
\begin{center}
    $\text{IF}\ x_1\ \text{is}\ \{L_{(1,1)}\ \text{or}\ L_{(1,2)}\}\ \text{and}\ x_2\ \text{is}\ \#\ \text{THEN}\ a\ \text{is}\ 2$
\end{center}
and can be encoded using \textit{GABIL encoding} \cite{de_jong_using_1993} as: $110 | 111 | 2$, where each clause of the antecedent is a binary mask, followed by the action to vote for, separated by vertical bars. In the inference engine of the FRBS, we use $f_\text{and} = \min$ for conjunction (ANDing) and $f_\text{or} = \max$ for disjunction (ORing) of membership values. Let $n$ be the number of rules in the RB. Given an input vector $\vec{x}$, a voting strength $g_a$ is calculated for each $a \in A$ via:
\begin{center}
    $g_a(\vec{x}) = \frac{\sum_{i=1}^{n} y_{(i,a)}\cdot\  \tau_{i}(\vec{x})}{\sum_{i=1}^{n}\tau_{i}(\vec{x})}$
\end{center}
where $y_{(i,a)}$ is the voting weight for action $a$ in the consequent of the $i^{th}$ rule, and $\tau_{i}(s)$ is the overall antecedent truth value (rule firing strength) of the $i^{th}$ rule in the context of $\vec{x}$; calculated through application of $f_\text{or}$ and $f_\text{and}$ to the membership values computed in the rule antecedent. The action to select is then determined via:
\begin{center}
    $\text{action} = \argmax\limits_{a \in A} g_a(\vec{x})$
\end{center}

\subsubsection{Measuring FRBS Complexity}
\label{sec:measureFRBSComplexity}
There are many possible ways to measure the complexity of an FRBS, including: number of rules in the RB, longest antecedent of any rule in the RB \cite{ishibuchi_multiobjective_2007}.
We choose an option that is based on the number of ``decision points'' represented in the system. For a $d$-dimensional feature space, let a \textit{fuzzy subspace} be defined as the intersection of $d$ fuzzy sets over the features: $\big(L_{(1,j)} \cap \ldots \cap L_{(d,j)}\big), i \in \{1, \ldots, d\}, j \in \{1, \ldots, m_i\}$. Such an intersection of fuzzy sets represents an \textit{elementary fuzzy rule} that is only capable of representing conjunctions, i.e. a single decision point in feature space. A CNF rule represents possibly many elementary rules, because disjunctions in such a rule represent generalisations over fuzzy subspaces. For an RB containing $n$ CNF rules, if the number of linguistic values specified in the $j^{th}$ clause of the $i^{th}$ rule's antecedent is given by $l_{(i,j)}$, then the total number of decision points embodied in the RB is:
\begin{equation}
    \text{complexity} = \sum_{i=1}^{n}\prod_{j=1}^{d}l_{(i,j)}
    \label{eqn:rbPhenotypicComplexityCalc}
\end{equation}
This is the measure of complexity that we use for an RB. The complexity of the overall FRBS is equal to the complexity of its RB.
\section{Related Work}
Many of the ideas required in this work have been considered previously in small combinations and in non-RL domains.
The cooperative coevolution architecture originally described in \cite{potter_cooperative_2000} has been adopted by GFSs to jointly evolve FRBS components, where one population (species) is dedicated to RBs and the other to DBs. For example, Fuzzy CoCo \cite{pena-reyes_fuzzy_2001} used this architecture to address the well-known classification problem of Wisconsin Breast Cancer Diagnosis. This particular system employed a fitness penalty for RB complexity, and so did not utilise multiobjectivity. However, it was able to evolve compact and interpretable FRBSs to address the problem, and it set a strong example for how CoCo could be used within a GFS.

A number of Pittsburgh GFSs have been designed to use MO mechanisms according to a survey conducted by Ishibuchi \cite{ishibuchi_multiobjective_2007}. An apposite example is the work of Ishibuchi et al. \cite{ishibuchi_three-objective_2001}, where an MO evolutionary algorithm evolves FRBSs to address various classification problems; finding trade-offs between three objective functions: i) maximise classification accuracy, ii) minimise the number of fuzzy rules, iii) minimise the total number of fuzzy rule antecedent conditions. These ideas need development to RL domains, especially adapting to credit assignment in multi-step problems. 

In the broader evolutionary computation context, MO and CoCo have also been combined in a single system, such as in the work of Iorio and Li \cite{iorio_cooperative_2004}. The validity of this system was demonstrated on a number of benchmark function optimisation problems, but not yet RL.
Peripherally related work includes Michigan style LCSs that use fuzzy logic rule representations, such as the Fuzzy Classifier System in \cite{valenzuela-rendon_reinforcement_1998} and Fuzzy-XCS in \cite{casillas_fuzzy-xcs_2007}. The former was applied to multi-step control problems (true RL), while the latter was only applied to single-step problems (function approximation and robot control). What is missing in all of these works is the combination of CoCo, MO, and FRBSs to address multi-step RL problems, and the intention of our work is to make a first attempt at addressing this gap.
\section{Mountain Car Environment}
\label{sec:mc}
In Mountain Car (MC), the agent must push a car out of a valley to the top of a mountain, as shown in the top plot of Figure~\ref{fig:mcBestFuzzySets}. State features are the position of the car on the horizontal axis: $x \in [-1.2, 0.5]$, and the horizontal component of the car's velocity: $\dot{x} \in [-0.07, 0.07]$. $A = \{1, 2\}$, representing push car 1: $\text{Left}$ or 2: $\text{Right}$. $R$ yields $-1$ at every time step, with $t_\text{max} = 200$. Discounting is not used (effectively $\gamma = 1$). The goal is reached when $x \ge 0.5$. Let $S_I = \{[-0.6, -0.4], 0\}$ with $Z$ constructed by sampling uniformly at random from $S_I$, such that the agent starts around the bottom of the valley with zero velocity. $\eta = 30$, with samples being drawn from $S_I$ using a fixed RNG seed, such that all performance evaluations use the same initial states.
These initial conditions make exploration difficult; if learning online (Michigan approach), the agent must somehow explore to the goal in order to learn how to escape the valley, then reinforce this path over time. In contrast, Pittsburgh approaches \textit{may find the task easier} if they are able to construct coherent policies that escape the valley, then improve them over successive generations.

\label{sec:envPerfBounds}
The minimum possible performance of a policy in MC is $-200$, indicating all $\eta$ rollouts were unable to reach the goal within $t_\text{max}$ steps. To calculate an upper bound on performance, we obtained a policy that was approximately optimal, and calculated the expected return achieved by it. To find this policy, we performed value iteration on a finely discretised version of MC (1000 bins per feature), yielding an approximately optimal action-value function $\widetilde{Q}$, then constructed an approximately optimal policy $\widetilde{\pi}$ by querying $\widetilde{Q}$ for each discretised state. The performance of $\widetilde{\pi}$ was $-96$ (rounded up to nearest integer).
\section{Cooperation and subspeciation}
\label{sec:coopSubspecies}
As previously mentioned, natural decomposition of an FRBS leads to the concept of a DB cooperating with an RB. We evolve FRBSs where both the DB and RB are subject to adaptation via a CoCo algorithm where one population represents DBs and the other RBs.
These populations are termed \textit{species} \cite{potter_cooperative_2000}. We choose to label the DB species as the \textit{first} population and the RB species as the \textit{second} population, and for the remainder of this section refer to them as $O_1$ and $O_2$ respectively.

In implementing this CoCo paradigm we must primarily ask: \textit{what kinds of structures are the two species searching over, and why?} A related secondary question is: \textit{how are individuals of each species genetically encoded?}
An answer to the primary question has to take into account the goal of the evolutionary process. In our case, we are performing an MO search over performance and complexity of FRBSs. Since the performance of an FRBS is governed by its interaction with the environment (and is outside of our control), diversity in this objective is a natural consequence of searching over many possible FRBSs of varying complexity. Therefore, there must be mechanisms built into the evolutionary process to support diversity in FRBS complexity. To achieve this, and to provide an answer to the primary question, we include a niching mechanism in our algorithm in the form of \textit{subspecies}. Since each population represents a species, a subspecies is a subpopulation.

To explain exactly what a subspecies is, and how it works in our CoCo paradigm, we have to begin to answer the secondary question posited above: \textit{how are individuals of each species genetically encoded?} Let $\text{idv}_1 \in O_1$ be a DB and $\text{idv}_2 \in O_2$ be an RB. In order to form a solution, $\text{idv}_1$ must cooperate with $\text{idv}_2$; however, it must actually be possible for these individuals to form a valid FRBS.
For example, assuming $d=2$, let there be an $\text{idv}_1$ for which $m_1 = m_2 = 2$, i.e. two fuzzy sets defined on each feature. Next, assume there is an $\text{idv}_2$ containing the following rule:
\begin{center}
$\text{IF}\ x_1\ \text{is}\ L_{(1,1)}\ \text{and}\ x_2\ \text{is}\ L_{(2,3)}\ \text{THEN}\ \ldots$
\end{center}
$\text{idv}_2$ does not make sense in the context of $\text{idv}_1$ because there is no fuzzy set $L_{(2,3)}$ defined in $\text{idv}_1$; therefore cooperation cannot occur. There must be a mechanism in the algorithm to prevent situations like this from occurring.

To accomplish this, each individual in both populations is assigned a (non-alterable) \textit{subspecies tag} $\sigma$ from a set of possible subspecies tags $\Sigma$, that indicates what subspecies it belongs to. A subspecies tag is a tuple of $d$ integers each $\ge 2$ representing the number of fuzzy sets defined on each of the $d$ feature domains (at least two on each). For example, the subspecies tag of $\text{idv}_1$ from the previous example is $(2,2)$. The subspecies tag determines the granularity of the fuzzy partitions over the feature space. Within each possible level of granularity, many levels of RB complexity are possible, as we explain in Section~\ref{sec:rbGeneticRepr}. For a given problem, this setup allows the evolutionary process to produce FRBSs with \textit{appropriate granularity and complexity} to address the problem. Diversity in granularity drives diversity in complexity, which enables diversity in performance; thus subspeciation is a critical mechanism in our MO search. Subspecies tags are primarily used to coordinate cooperation between individuals of both species. Let $\text{idv}.\sigma$ denote the subspecies tag of $\text{idv}$. Cooperation is \textit{restricted to be performed intra-subspecies}, such that $\text{idv}_1$ and $\text{idv}_2$ can only cooperate if $\text{idv}_1.\sigma = \text{idv}_2.\sigma$.

To genetically represent individuals of a given subspecies, we use a \textit{fixed-length positional vector encoding}, where the length of an individual's genotype is dependent on its subspecies tag. For individuals in $O_1$, the genotype encodes ``reference coordinates'' along each feature domain that are used to construct fuzzy sets, as described in Section~\ref{sec:lvGeneticRepr}.
For individuals in $O_2$, the genotype encodes the advocation of actions in fuzzy subspaces, as detailed in Section~\ref{sec:rbGeneticRepr}.

\subsection{DB Genetic Representation}
\label{sec:lvGeneticRepr}
For an individual $\text{idv} \in O_1$, each element $\sigma_i$ of $\text{idv}.\sigma$ represents the number of fuzzy sets used to partition feature $i$. To encode $m = \sigma_i$ fuzzy sets, $m$ values are required, as explained below. Therefore, the length of such an individual's genotype is given by:
\begin{equation}
    \lambda_1(\sigma) = \sum\limits_{i=1}^{d} \sigma_i
    \label{eqn:genotype-len-lv}
\end{equation}

\begin{figure}[h]
    \centering
    \includegraphics[trim={0 0.25cm 0.7cm 0.1cm},clip,width=\columnwidth]{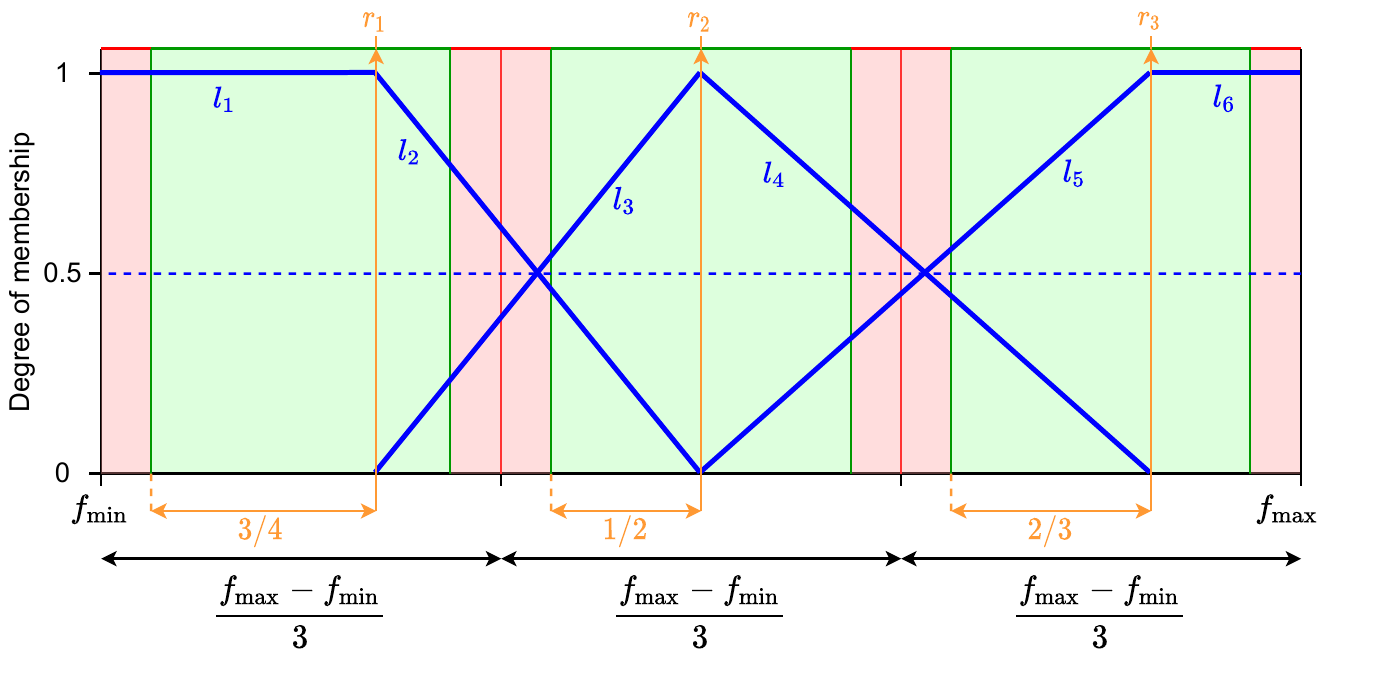}
    \caption{Example decoding of alleles from a DB individual over a single feature domain to produce fuzzy sets.}
    \label{fig:lvGenotypeDecoding}
\end{figure}

Since the genotype is a vector, its genes can be logically split into sections that are responsible for a specific feature.  Alleles in the genotype are real numbers in the range $[0, 1]$. Figure~\ref{fig:lvGenotypeDecoding} depicts an example of genotype decoding for an individual in $O_1$, on a single feature domain $[f_\text{min}, f_\text{max}]$. This process is repeated with the appropriate alleles for each feature domain to create fuzzy sets for all linguistic variables. Applicable alleles in this example are $(\frac{3}{4}, \frac{1}{2}, \frac{2}{3})$, shown in orange at the bottom of the figure. Since there are three alleles, three fuzzy sets are constructed along the domain; $m = 3$. The outer two fuzzy sets on the extremes of the domain are trapezoidal in shape and the inner fuzzy set is triangular in shape. In general, for $m = 2$ there is no inner triangular fuzzy set and for $m \ge 4$, there are multiple inner triangular fuzzy sets.

First, the domain is split into $m$ equal width subdomains. Next, a fraction $\omega$ of the center of each subdomain is marked as a \textit{valid} region (shaded green, red areas are \textit{invalid}). We set $\omega = 0.75$, i.e. 75\% of the middle of each subdomain is valid, 12.5\% on each side (remaining 25\%) is invalid. Each allele specifies a relative fraction \textit{of the width of the valid region --- measuring from the left hand side of the region} --- at which a ``reference coordinate'' is placed along the domain. $m$ reference coordinates $r_1, \ldots, r_m$ are placed, shown in orange at the top of the figure. 

Using these reference coordinates, lines are drawn to construct the fuzzy sets. To construct the outer two trapezoidal fuzzy sets, lines are drawn between the following pairs of points:
\begin{center}
    $\big( (f_\text{min}, 1), (r_1, 1) \big),\ \big( (r_1, 1), (r_2, 0) \big),\ \big( (r_{m-1}, 0), (r_m, 1) \big),$\\ $\big( (r_m, 1), (f_\text{max}, 1) \big)$
\end{center}
which correspond to lines $l_1, l_2, l_5, l_6$ in our example. Next, lines for the inner triangular membership functions are created: for $r_i, i \in \{2, \ldots, m-1\}$, lines are drawn from the point $(r_i, 1)$ to points $(r_{i-1}, 0)$ (positive gradient) and $(r_{i+1}, 0)$ (negative gradient), corresponding to lines $l_3, l_4$ in our example. Because of the concept of valid/invalid regions, this construction process has the desirable property of ensuring that there is a minimum amount of separation between neighbouring fuzzy sets, which reduces overlap and aids linguistic distinguishability \cite{cordon_genetic_2001}.

\subsection{RB Genetic Representation}
\label{sec:rbGeneticRepr}
An individual $\text{idv}_2 \in O_2$ must operate in the context of fuzzy sets specified by an individual $\text{idv}_1 \in O_1$. Because of this, there are only a certain number of fuzzy subspaces in which $\text{idv}_2$ can advocate actions. The number of fuzzy subspaces is the number of possible fuzzy set intersections, i.e. the product of the number of fuzzy sets defined on each feature dimension. The length of $\text{idv}_2$'s genotype is given by this number:
\begin{equation}
    \lambda_2(\sigma) = \prod\limits_{i=1}^{d} \sigma_i
    \label{eqn:genotype-len-rb}
\end{equation}
Expressed as a vector, the genes in the genotype are ordered in the same order as nested for-loops over the fuzzy intersections, e.g. for $d = 2, m_1 = 3, m_2 = 2$, the genotype is of length six with genes specifying the following fuzzy subspaces:
\noindent
\begin{center}
    $\big(L_{(1,1)}\cap\ L_{(2,1)}\big),\ \big(L_{(1,1)}\cap\ L_{(2,2)}\big),\ \big(L_{(1,2)}\cap\ L_{(2,1)}\big),$ \\ $\big(L_{(1,2)}\cap\ L_{(2,2)}\big),\ \big(L_{(1,3)}\cap\ L_{(2,1)}\big),\ \big(L_{(1,3)}\cap\ L_{(2,2)}\big)$
\end{center}
Each gene has alleles from the set: $A \cup \{0\}$, which select the action to advocate in the fuzzy subspace. The alleles from $A$ are self-explanatory, but the $0$ allele signifies that \textit{no} action is specified, the fuzzy subspace is ignored. Thus, it is possible for an RB genotype to be \textit{under-specified} in that actions do not have to be advocated in all fuzzy subspaces. Subspaces with a $0$ allele are said to be \textit{unspecified}, else they are \textit{specified}. The possible underspecification of an RB is how different levels of complexity are achieved. This has important implications for measuring the performance of an FRBS, as we expand on in Section~\ref{sec:alg}.

This genetic encoding represents a set of elementary fuzzy rules that only allow conjunction of fuzzy sets (c.f. Section~\ref{sec:measureFRBSComplexity}). However, since we actually use CNF fuzzy rules in the RB phenotype, there is a mechanism to \textit{merge fuzzy rules together} during genotype decoding in order to create CNF rules where commonalities are ``factored out''. A merge creates a disjunction that generalises over multiple fuzzy subspaces, and occurs when the binary encodings of two rules share the same bits in \textit{all but one} clause --- the differing clause creates the disjunction. This process is very similar to how a Karnaugh map is constructed for simplifying Boolean expressions. For example, using the same format of genotype as above and given the following specification of linguistic variables and their corresponding linguistic values:
\begin{center}
    $x_1: \{L_{(1,1)}: \text{L},\ L_{(1,2)}: \text{M},\ L_{(1,3)}: \text{H}\},\
    x_2: \{L_{(2,1)}: \text{L},\ L_{(2,2)}: \text{H}\}$
\end{center}

\begin{figure}[h]
    \centering
    \includegraphics[width=\columnwidth]{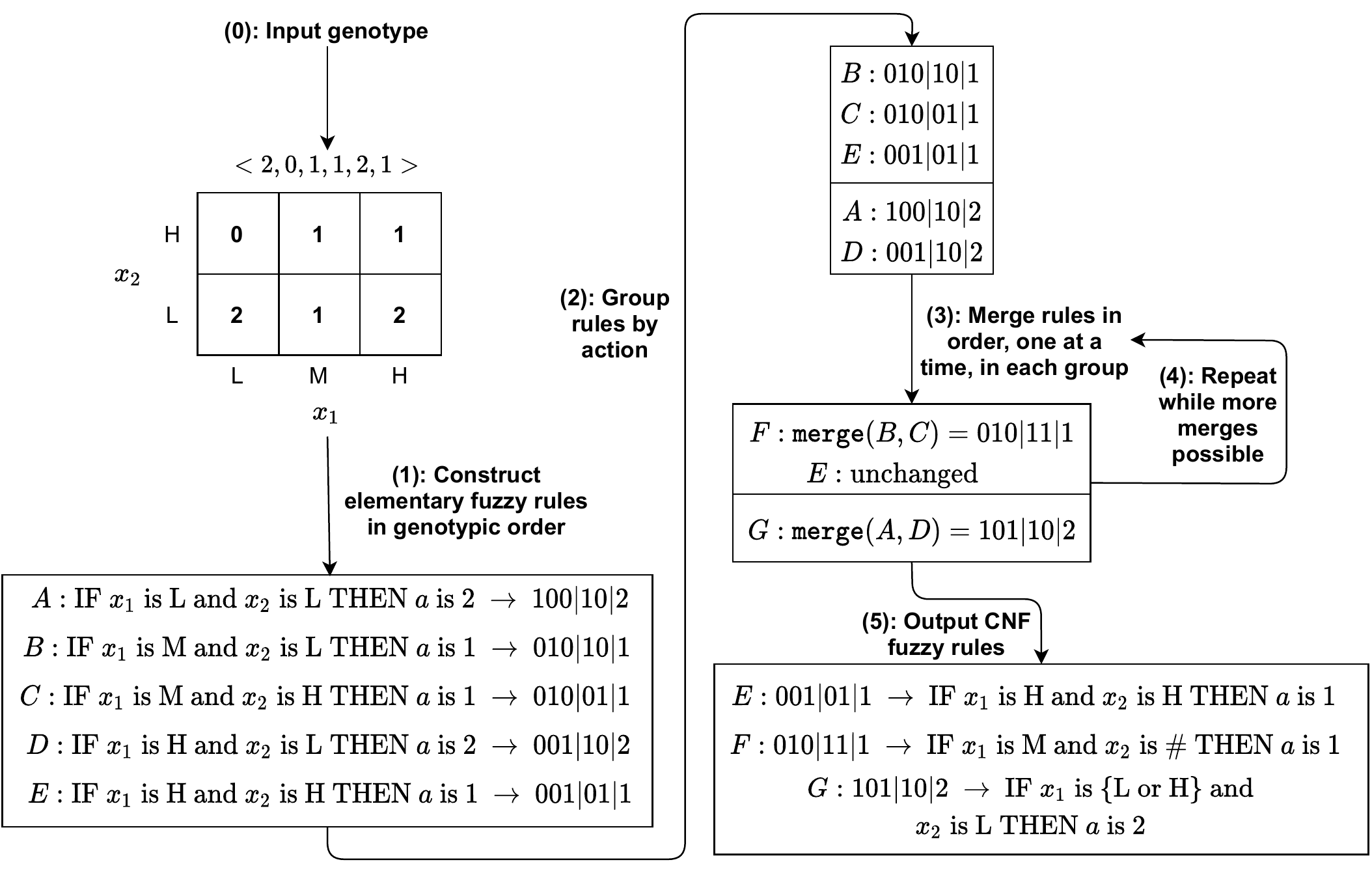}
    \caption{Example decoding of an RB genotype to produce a phenotype of CNF fuzzy rules.}
    \label{fig:cnfConstruction}
\end{figure}

Figure~\ref{fig:cnfConstruction} depicts how the genotype $<2,0,1,1,2,1>$ is decoded into a phenotype of CNF fuzzy rules.
Note that in this example, only one application of step $(3)$ is necessary, but in general step $(3)$ may need to be repeated multiple times. Also note that the order of merging in step $(3)$ is fixed; given our example rule $B$ would always be merged with rule $C$, and it is not possible for rule $C$ to be merged with rule $E$ (even though this would result in an equally valid CNF rule).

Why use this encoding; why not just encode CNF rules directly? There are two good reasons: it is impossible to have rules that are \textit{over-specified} in the sense of being either i) redundant, or ii) contradictory, since \textit{at most} one action can be specified in every fuzzy subspace. Redundancy occurs when two rules $A$ and $B$ share the same consequent but rule $A$ has an antecedent that logically subsumes $B$'s. Contradictions occur when $A$ and $B$ specify common fuzzy subspace(s) in their antecedents but their consequents differ \cite{cordon_genetic_2001}. Both of these situations are problematic, contradictions more so than redundancies. There has been much attention in the literature on how to deal with these situations; some approaches allow them to genetically manifest and phenotypically deal with them via a conflict resolution procedure when evaluating the rule set, others employ corrections in genotype space to remove them if they occur \cite{casillas_consistent_2007}. We take the approach of making them unable to occur.

\subsection{Complexity Bounds}
\label{sec:complexityBounds}
In Section~\ref{sec:measureFRBSComplexity} we defined how the complexity of an RB of CNF rules is (phenotypically) calculated. However, given the RB genotype to phenotype decoding example shown in Figure~\ref{fig:cnfConstruction}, it is apparent that RB complexity can be measured genotypically as the \textit{number of specified alleles} in the RB genotype. In the example, the number of specified alleles in the genotype is $5$, which is exactly the complexity of the RB of CNF rules: applying Equation~\ref{eqn:rbPhenotypicComplexityCalc} gives $1\times1=1$ for rule E, $1\times1=1$ for rule F, $2\times1=2$ for rule G, $1+1+2=5$ total.
Therefore we actually measure RB complexity genotypically.

We use the following bounds for complexity: minimum complexity for any RB genotype is equal to $k$ (number of actions in $A$), meaning we give each RB the opportunity to advocate at least one rule for each possible action. Maximum complexity is equal to the \textit{maximum possible number of specified alleles for any RB genotype}, equivalent to the length of the longest genotype of any RB subspecies.
\section{Fuzzy MoCoCo}
\label{sec:alg}
We now present our algorithm for performing multiobjective cooperative coevolution of FRBSs. Our algorithm most closely resembles the system used in \cite{iorio_cooperative_2004}, in that it uses the same overarching framework for integrating the cooperative coevolutionary mechanisms from CCGA \cite{potter_cooperative_2000} with the Pareto multiobjective and elitist features of NSGA-II \cite{deb_fast_2002}. Algorithm~\ref{alg:fuzzyMOCOCO} presents a top-level overview of our algorithm: Fuzzy MoCoCo (\textbf{M}ulti\textbf{o}bjective \textbf{Co}operative \textbf{Co}evolution). Algorithms \ref{alg:buildCollabrMap}--\ref{alg:breedChildren} detail the main functions used in Algorithm~\ref{alg:fuzzyMOCOCO}.
In these algorithms and for the remainder of this paper, the following notation is used:
\begin{itemize}
    \item $\delta$ --- a probability mass function (PMF) over subspecies tags
    \item $P$ --- a parent population
    \item $Q$ --- a child population
    \item $R$ --- a combined parent \textit{and} child population
    \item $O$ --- a population: used in a general sense when \textit{any} of ($P, Q, R$) could be expected
    \item $O_i, i \in \{1,2\}$ --- the $i^{th}$ population
    \item $O^{\sigma}, \sigma \in \Sigma$ --- a subpopulation; individuals in $O$ with subspecies tag $\sigma$
    \item $O_i^{\sigma}$ --- a subpopulation of the $i^{th}$ population
    \item $S$ --- a set of solutions (FRBSs)
    \item $\prec_\text{cc}$ -- NSGA-II crowded comparison operator; partial ordering based on (Pareto front rank, crowding distance) pairs
\end{itemize}

\begin{algorithm}
    \KwIn{$\mathcal{E}$, $\Sigma$}
    \DontPrintSemicolon
    \SetAlgoLined
    $\delta_1, \delta_2 = \texttt{makeSubspeciesDists}(\Sigma)$\;\label{func:makeSubspeciesDists}
    $P_1 = \texttt{initDBPop}(\delta_1)$\;\label{func:initDBPop}
    $P_2 = \texttt{initRBPop}(\delta_2)$\;\label{func:initRBPop}
    $Q_1 = \emptyset$\;
    $Q_2 = \emptyset$\;
    $\textit{gen} = 0$\;
    \While{$\textit{gen} < \textit{numGens}$}{
        $\chi = \texttt{buildCollabrMap}(P_1, P_2, \Sigma, gen)$\;\label{alg:fuzzyMOCOCO-evalStart}
        \If{$ \textit{gen} == 0$}{
            $S = \texttt{buildSolnSet}(P_1, P_2, \Sigma, \chi)$\;
            $\texttt{evalSolnSet}(S, \mathcal{E})$\;
            $\texttt{assignIndivsCredit}(P_1, S)$\;
            $\texttt{assignIndivsCredit}(P_2, S)$\;
        }\Else{
            $S = \texttt{buildSolnSet}(Q_1, Q_2, \Sigma, \chi)$\;
            $\texttt{evalSolnSet}(S, \mathcal{E})$\;
            $\texttt{assignIndivsCredit}(Q_1, S)$\;
            $\texttt{assignIndivsCredit}(Q_2, S)$\;\label{alg:fuzzyMOCOCO-evalEnd}
        }
        $R_1 = P_1 \cup Q_1$\;\label{alg:fuzzyMOCOCO-reprodStart}
        $R_2 = P_2 \cup Q_2$\;
        $P_1 = \texttt{archiveParentPop}(R_1, \delta_1, \vert P_1 \vert)$\;
        $P_2 = \texttt{archiveParentPop}(R_2, \delta_2, \vert P_2 \vert)$\;
        $Q_1 = \texttt{breedChildren}(P_1, \delta_1)$\;
        $Q_2 = \texttt{breedChildren}(P_2, \delta_2)$\;\label{alg:fuzzyMOCOCO-reprodEnd}
        $\textit{gen} = \textit{gen}+1$\;
    }
    \Return{$(R_1, R_2, S)$}
    \caption{\texttt{Fuzzy MoCoCo}}
    \label{alg:fuzzyMOCOCO}
\end{algorithm}
\vspace{-0.25cm}
\begin{algorithm}
    \DontPrintSemicolon
    \SetAlgoLined
    \KwIn{$P_1, P_2, \Sigma, \textit{gen}$}
    $\chi = $ empty mapping of subpop specification pairs to indivs\;
    \For{$(i, \sigma) \in (\{1, 2\} \times \Sigma)$}{
        $\chi[(i,\sigma)] = \texttt{selectCollabrs}(P_i^{\sigma}, \textit{gen})$\;\label{func:selectCollabrs}
    }
    \Return{$\chi$}
    \caption{\texttt{buildCollabrMap}}
    \label{alg:buildCollabrMap}
\end{algorithm}

\begin{algorithm}
    \DontPrintSemicolon
    \SetAlgoLined
    \KwIn{$O_1, O_2, \Sigma, \chi$}
    $S = \emptyset$\;
    $\textit{popNums} = \{1, 2\}$\;
     \For{$(i, \sigma) \in (\textit{popNums} \times \Sigma)$}{
        $\textit{subpop} = O_i^{\sigma}$\;
        $j = $ opposite of $i$ in $\textit{popNums}$\;
        $\textit{collbars} = \chi[(j,\sigma)]$\;
        \For{$(\textit{idv}, \textit{collabr}) \in (\textit{subpop} \times \textit{collabrs})$}{
            $\textit{soln} = \texttt{makeFRBS}(\textit{idv}, \textit{collabr})$\;\label{alg:buildSolnSet-makeSoln}
            $S = S \cup \{\textit{soln}\}$\;
        }
    }
    \Return{$S$}
    \caption{\texttt{buildSolnSet}}
    \label{alg:buildSolnSet}
\end{algorithm}

\begin{algorithm}
    \DontPrintSemicolon
    \SetAlgoLined
    \KwIn{$S, \mathcal{E}$}
    \For{$\textit{soln} \in S$}{
        $\textit{soln.perf} = \texttt{calcPerformance}(\textit{soln},\mathcal{E})$\;
        $\textit{soln.comp} = \texttt{calcComplexity}(\textit{soln})$\;
    }
    $\texttt{assignParetoFrontRanks}(S)$\;\label{alg:evalSolnSet-pfrs}
    $\texttt{assignCrowdingDists}(S)$\;\label{alg:evalSolnSet-crowdDists}
    \caption{\texttt{evalSolnSet}}
    \label{alg:evalSolnSet}
\end{algorithm}

\begin{algorithm}
    \DontPrintSemicolon
    \SetAlgoLined
    \KwIn{$O, S$}
    \For{$\textit{idv} \in O$}{
        $C = $ set of solns in $S$ that contain $\textit{idv}$ as a component\;
        $C' = \texttt{crowdedComparisonSort}(C)$\;
        $\textit{best} = $ first soln in $C'$\;
        $\textit{idv.perf} = \textit{best.perf}$\;
        $\textit{idv.comp} = \textit{best.comp}$\;
    }
    \caption{\texttt{assignIndivsCredit}}
    \label{alg:assignIndivsCredit}
\end{algorithm}

\begin{algorithm}
    \DontPrintSemicolon
    \SetAlgoLined
    \KwIn{$R, \delta, \textit{numParents}$}
    $R' = $ copy of $R$\;
    $P = \emptyset$\;
    \While{$\vert P \vert < \textit{numParents}$}{
        $\sigma = $ draw sample from $\delta$\;\label{alg:archiveParentPop-whileStart}
        $\textit{subpop} = R'^{\sigma}$\;
        \If{$\textit{subpop}$ not empty}{
            $C = \texttt{crowdedComparisonSort}(\textit{subpop})$\;
            $\textit{best} = $ first soln in $C$\;
            $R' = R' - {\textit{best}}$\;
            $P = P \cup \{\textit{best}\}$\;\label{alg:archiveParentPop-whileEnd}
        }
    }
    \Return{$P$}
    \caption{\texttt{archiveParentPop}}
    \label{alg:archiveParentPop}
\end{algorithm}

\begin{algorithm}
    \DontPrintSemicolon
    \SetAlgoLined
    \KwIn{$P, \delta$}
    $Q = \emptyset$\;
    \While{$\vert Q \vert < \vert P \vert$}{
        $\sigma = $ draw sample from $\delta$\;
        $\textit{subpop} = P^{\sigma}$\;
        $\textit{parentA} = \texttt{selection}(\textit{subpop})$\;
        $\textit{parentB} = \texttt{selection}(\textit{subpop})$\;
        $\textit{childA, childB} = \texttt{crossoverMutate}(\textit{parentA, parentB})$\;
        $Q = Q \cup \{\textit{childA, childB}\}$\;
    }
    \Return{$Q$}
    \caption{\texttt{breedChildren}}
    \label{alg:breedChildren}
\end{algorithm}

As input to Algorithm~\ref{alg:fuzzyMOCOCO}, $\mathcal{E}$ and $\Sigma$ must be specified. The main generational loop of Algorithm~\ref{alg:fuzzyMOCOCO} can be split into two phases: the top half (lines \ref{alg:fuzzyMOCOCO-evalStart}--\ref{alg:fuzzyMOCOCO-evalEnd}, evaluation phase) builds solutions via cooperation, evaluates these solutions in the MO space, then assigns credit (objective values) to individuals based on their participation.
The second half (lines \ref{alg:fuzzyMOCOCO-reprodStart}--\ref{alg:fuzzyMOCOCO-reprodEnd}, reproductive phase) archives the best solutions found so far, then breeds new child populations from these archives.

The reproductive phase is the same for every iteration of the loop, but the evaluation phase has different behaviour for the first vs. subsequent generations.
In the first generation ($\textit{gen} = 0$), $Q_1$ and $Q_2$ are empty and so $P_1$ must cooperate with $P_2$. This is a bootstrapping generation, where the initial parents are evaluated. In every subsequent generation, $P_1$ cooperates with $Q_2$ and $P_2$ cooperates with $Q_1$, with the purpose of evaluating individuals in $Q_1$ and $Q_2$. This means each individual is evaluated exactly once: in the first generation for the initial parents, and in the generation after its conception for every child. Elitism appears in the form of archiving individuals from the combined populations $R_1$ and $R_2$ as parents.
The function \texttt{makeSubspeciesDists} (Algorithm~\ref{alg:fuzzyMOCOCO}, line~\ref{func:makeSubspeciesDists}) creates a subspecies PMF for both populations: $\delta_1$ and $\delta_2$. These PMFs specify what fraction of the search should be (probabilistically) dedicated to each subspecies: subspecies with larger search spaces receive a larger fraction of the available resources. For $i \in \{1, 2\}, \sigma \in \Sigma$, $\delta_i$ is initialised as:
\begin{equation*}
    \delta_i[\sigma] = \frac{\beta^{\lambda_i(\sigma)}}{\sum_{\sigma' \in \Sigma}\beta^{\lambda_i(\sigma')}}
    \label{eqn:pi-calc}
\end{equation*}
where $\lambda_i(\sigma)$ calculates the length of the genotype used by subspecies $\sigma$ in population $i$ (Equation~\ref{eqn:genotype-len-lv}, Equation~\
\ref{eqn:genotype-len-rb}). $\beta \ge 1$ is a hyperparameter that controls disparity in probability mass allocation, larger values of $\beta$ allocate more mass to subspecies with longer genotypes.
$P_1$ and $P_2$ are initialised in the functions \texttt{initDBPop} and \texttt{initRBPop}. These functions create $\textit{dbPopSize}$ and $\textit{rbPopSize}$ individuals respectively, determining which subspecies to create by drawing samples from $\delta_1$ and $\delta_2$, respectively. \texttt{initDBPop} initialises alleles randomly from $\mathcal{U}(0, 1)$. For \texttt{initRBPop}, a hyperparameter $\textit{rbPUnspec}$ controls the probability of initialising an allele as unspecified. The remaining alleles ($a \in A$) each have an initialisation probability of $\frac{1 - \textit{rbPUnspec}}{k}$.

Algorithms \ref{alg:buildCollabrMap} and \ref{alg:buildSolnSet} implement cooperation between individuals as previously described. To select collaborators for cooperation, the function \texttt{selectCollbars} (Algorithm~\ref{alg:buildCollabrMap}, line~\ref{func:selectCollabrs}) has two different behaviours, depending on the generation counter. During the first generation there is no information about the objective values of any individual, so Pareto front ranks and crowding distances cannot be computed, and $\prec_{cc}$ cannot be applied. Therefore \textit{two} collaborators are randomly selected in each subpopulation. During subsequent generations, again \textit{two} collaborators are selected from each subpopulation, but are taken as: i) the best individual according to $\prec_\text{cc}$, ii) a random individual from the remainder of the subpopulation.

In Algorithm~\ref{alg:evalSolnSet}, the performance and complexity of solutions in $S$ is evaluated. Because we allow RBs to be underspecified (see Section~\ref{sec:rbGeneticRepr}), it is possible that an FRBS can \textit{fail} its performance evaluation, i.e. an input state is reached that is not covered by any rule. This is an inherent disadvantage of a Pittsburgh system as opposed to a Michigan system, the latter makes this impossible via a covering mechanism. If such a scenario is encountered, the FRBS is assigned \textit{a performance equal to the environmental lower bound} (see Section~\ref{sec:envPerfBounds}). Evaluation of complexity is done as per Section~\ref{sec:complexityBounds}.

Following performance and complexity evaluation, the function \texttt{assignParetoFrontRanks} determines the Pareto front ranks of solutions in $S$, implemented as the same fast non-dominated sort from NSGA-II. A Pareto front of rank $i$ is denoted by $\mathcal{F}_i$, $\mathcal{F}_1$ representing the set of non-dominated solutions on the frontier of the search.
Finally, \texttt{assignCrowdingDists} is used to determine the crowding distance of solutions, again in the same fashion as NSGA-II. To do this, lower and upper bounds for both performance and complexity must be known. Performance bounds are a property of $\mathcal{E}$ (see Section~\ref{sec:envPerfBounds}). Complexity bounds are as discussed in Section~\ref{sec:complexityBounds}.

Algorithm~\ref{alg:assignIndivsCredit} assigns credit to individuals in a population according to the \textit{best} solution they participated in. The notion of best is determined via application of $\prec_{cc}$ (\texttt{crowdedComparisonSort} function). Algorithm~\ref{alg:archiveParentPop} performs an NSGA-II style archiving procedure, selecting a new $P$. The distribution of subspecies tags in $P$ is reflective of $\delta$, in that the main loop (lines \ref{alg:archiveParentPop-whileStart}--\ref{alg:archiveParentPop-whileEnd}) firstly draws a subspecies tag from $\delta$. Then, the best individual from the corresponding subpopulation is archived in $P$, and removed from the set of candidate parents $R'$. This is repeated until $P$ is full.

Algorithm~\ref{alg:breedChildren} generates a new child population $Q$, via application of a GA on $P$. The selection operator is tournament selection with a tournament size of 2, using $\prec_\text{cc}$ to rank individuals. Like Algorithm~\ref{alg:archiveParentPop}, $\delta$ is sampled to preserve the distribution of subspecies tags in $Q$. Reproduction is done \textit{intra-subspecies}: the subspecies tag drawn from $\delta$ determines the subpopulation to select parents from. We use a real-coded GA on $P_1$ (since its genotypes are vectors of real numbers), with crossover implemented as line recombination \cite{luke_essentials_2013}, performed with probability $\textit{dbPCross}$. Mutation is Gaussian noise, zero mean, standard deviation controlled by $\textit{dbMutSigma}$.

For $P_2$ we use uniform crossover, performed with probability $\textit{rbPCross}$ per allele. Mutation allows each allele (being one of $k+1$ values from $A \cup \{0\}$) to switch to one of the other $k$ alleles, each with probability $\frac{1}{k}$. The probability of performing such a mutation is $\textit{rbPMut}$ per allele. Crossover and mutation probabilities are deliberately constant across subspecies, inciting more exploration (more frequent crossover and mutation) in subspecies with longer genotypes.
Because we specify a minimum RB complexity, we include a \textit{repair} operator, applied after mutation in $Q_2$. This operator rectifies situations where the number of specified alleles in an RB genotype is less than the minimum complexity --- altering the genotype to be \textit{of minimum complexity} by randomly selecting an appropriate number of unspecified alleles and assigning them random values from $A$.
\section{Results}
We conducted thirty independent runs of Fuzzy MoCoCo\footnote{Source code for algorithm available at: \url{https://github.com/jtbish/fuzzy-mococo}} on the OpenAI Gym implementation of MC\footnote{\url{https://gym.openai.com/envs/MountainCar-v0/}}, in order to determine if parsimonious, high-performing policies could be found. Hyperparameters for each run were:
$\textit{rbPUnspec}{=}0.1, \textit{numGens}{=}50, \textit{dbPCross}{=}0.75,$\\ $\textit{dbMutSigma}{=}0.02, \textit{rbPCross}{=}0.25, \textit{rbPMut}{=}0.05, \eta{=}30,$\\ $\Sigma{=}\{(2,2),(3,3),(4,4),(5,5)\}, \beta{=}1.125, \textit{dbPopSize}{=}300,$\\ $\textit{rbPopSize}{=}600$. 
$\beta$, $\textit{dbPopSize}$, and $\textit{rbPopSize}$ were set in tandem to allocate each subspecies approximately ten times as many individuals as the dimensionality of its search space.
\begin{figure}
    \centering
    \includegraphics[trim={0.1cm 0.25cm 0.1cm 0.1cm},clip,width=\columnwidth]{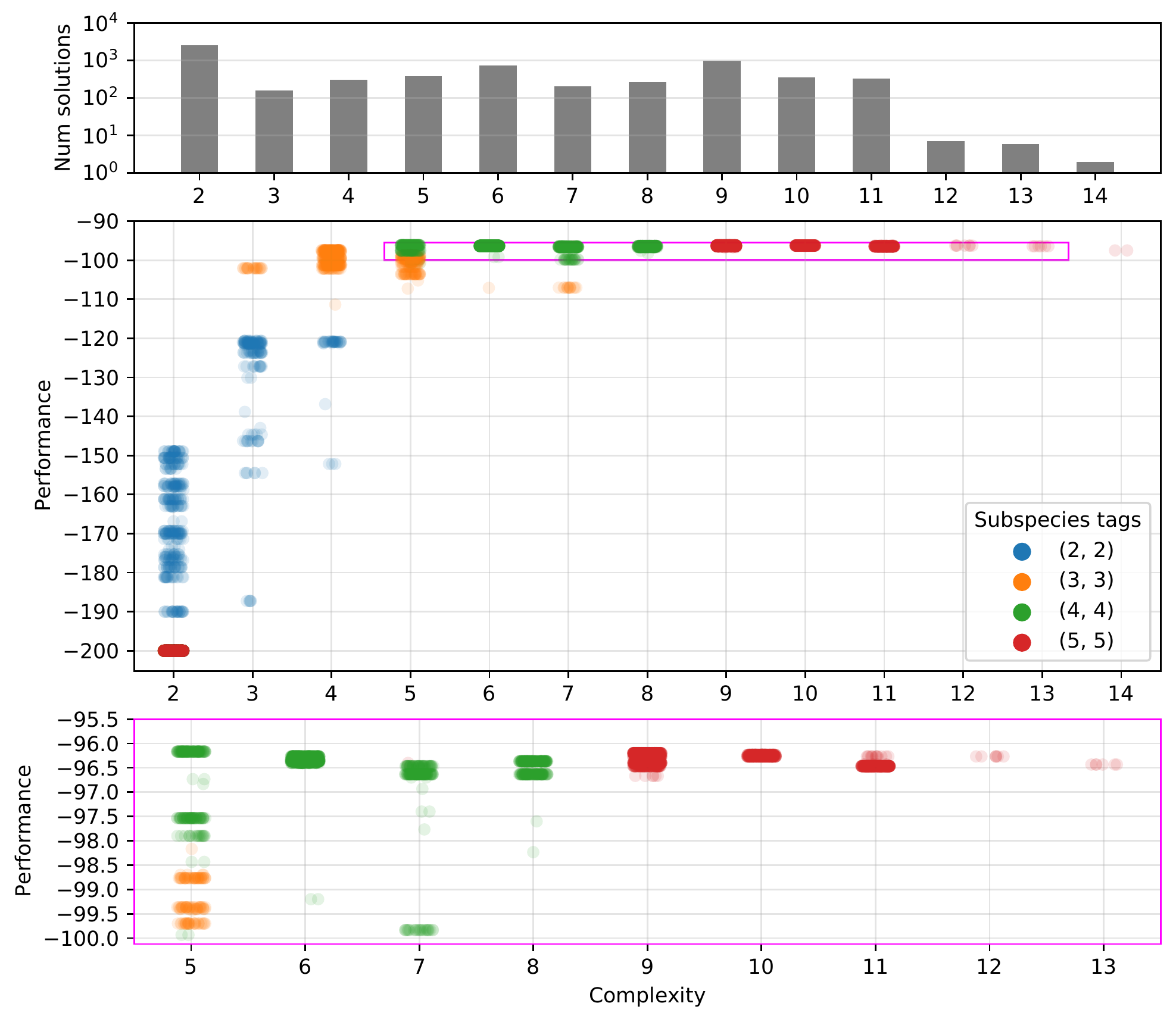}
    \caption{Middle: scatter plot of merged $\mathcal{F}_1$s over thirty Fuzzy MoCoCo runs on MC. Bottom: zoomed in view of highlighted region in middle plot. Top: histogram of complexity values in middle plot.}
    \label{fig:mcParetoFronts}
\end{figure}
Figure~\ref{fig:mcParetoFronts} (middle) shows a scatter plot of $\mathcal{F}_1$ yielded by each run (thirty $\mathcal{F}_1$s, plotted on one set of axes). Each solution is plotted as an individual point with high transparency, giving some indication of solution density in a particular area. A small amount of jitter is used on the complexity axis to make individual points in low density areas more visible. Points are coloured by their subspecies tag. Note that in areas of very high density (e.g. around $(2, -200)$), multiple points are plotted on top p of one another and so cannot be distinguished; this is a limitation of the plotting technique. The top plot of the figure shows the frequency of each level of complexity. The bottom plot of the figure shows a zoomed in view of the magenta area in the middle plot.
From this figure, we observe the following:
\begin{enumerate}
    \item A large number of solutions are of minimum complexity, minimum performance: $(2, -200)$.\label{mcPointOne}
    \item Increasing complexity up to 5 provides increased performance, but from then on provides no improvement.\label{mcPointTwo}
    \item Solutions offering the best tradeoff between performance and complexity are of complexity 5, subspecies $(4,4)$.\label{mcPointThree}
\end{enumerate}
The first observation manifests because the minimum performance attainable in MC is $-200$, indicating that no performance rollouts were able to reach the goal. This can easily occur, via an FRBS that either i) has rules that do not cooperate enough to reach the goal, or ii) is too underspecified and so fails its performance evaluation.

We chose one of the solutions from the complexity 5 subspecies $(4,4)$ group as the \textit{best} solution found by any run. The performance of this best solution was $-96.17$, while the performance of our approximately optimal policy was $-96$. This is notable: \textit{our algorithm has produced a policy that has almost attained the upper bound of performance}.
Due to limited available space, a discussion of the performance and computational complexity of Fuzzy MoCoCo vs. other learning approaches is omitted. Such a discussion is a task for future work.

\begin{figure}
    \centering
    \includegraphics[trim={0 0.33cm 0.1cm 0},clip,width=\columnwidth]{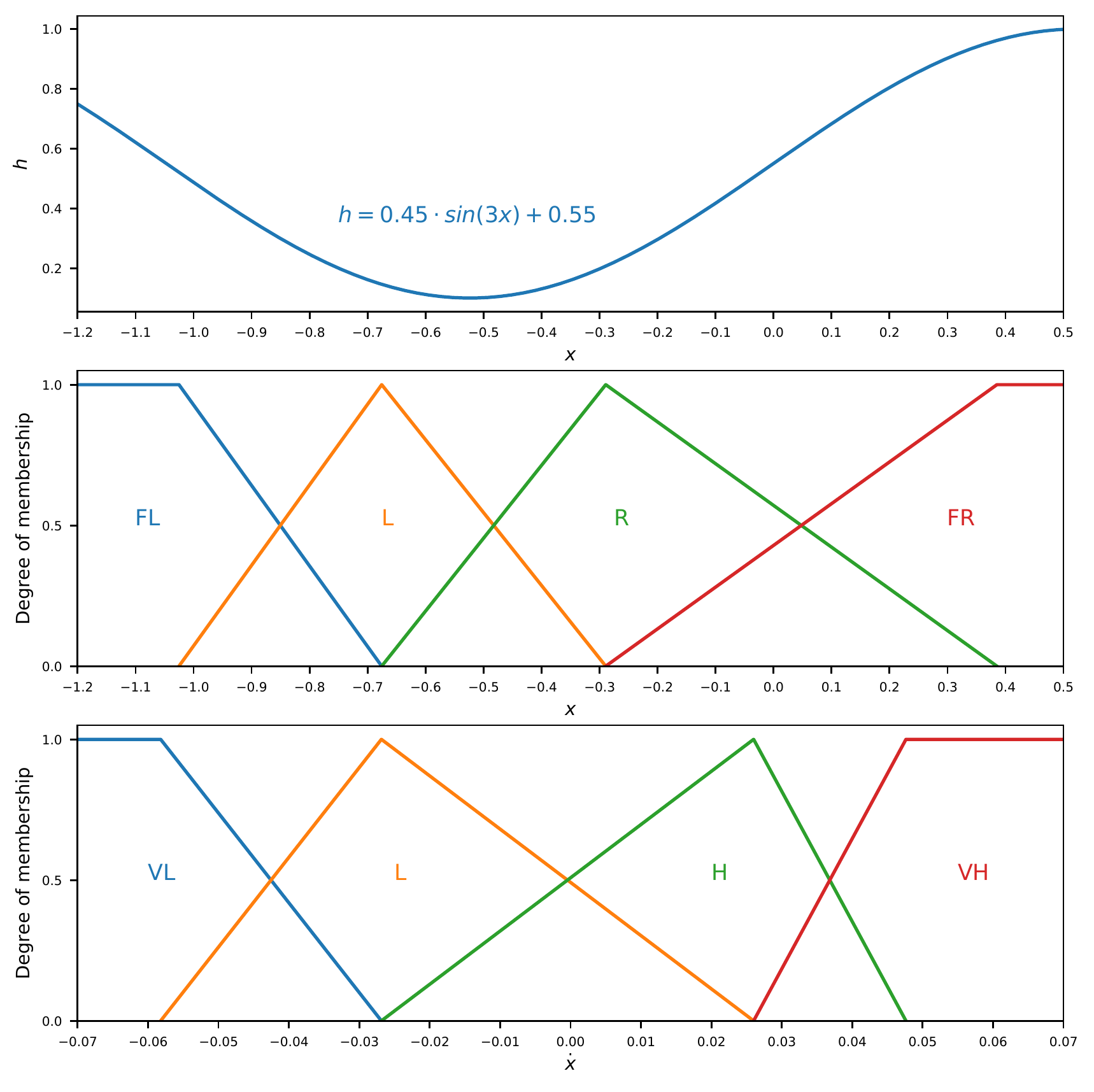}
    \caption{Top: curvature of the MC valley. Middle, bottom: fuzzy sets used by the best evolved FRBS in MC.}
    \label{fig:mcBestFuzzySets}
\end{figure}

Figure~\ref{fig:mcBestFuzzySets} shows the fuzzy sets used by the best solution (middle, bottom), accompanied by the curvature of the valley (top) to make fuzzy sets over $x$ easier to interpret. Linguistic values are as follows:
$x: \{\text{FL: Far\ Left}, \text{L: Left}, \text{R: Right}, \text{FR: Far\ Right}\}, \dot{x}: \{\text{VL: Very\ Low}, \text{L: Low}, \text{H: High}, \text{VH: Very\ High}\}$.
The RB used by the best solution is:
\begin{enumerate}[label=(\Alph*)]
    \item $\text{IF}\ x\ \text{is}\ \text{L}\ \text{and}\ \dot{x}\ \text{is}\ \text{L}\ \text{THEN}\ a\ \text{is}\ 1\ \text{(Left)}$\label{mcBestRuleA}
    \item $\text{IF}\ x\ \text{is}\ \{\text{FL}\ \text{or}\ \text{L}\ \text{or}\ \text{FR}\}\ \text{and}\ \dot{x}\ \text{is}\ \text{H}\ \text{THEN}\ a\ \text{is}\ 2\ \text{(Right)}$\label{mcBestRuleB}
    \item $\text{IF}\ x\ \text{is}\ \text{R}\ \text{and}\ \dot{x}\ \text{is}\ \text{VH}\ \text{THEN}\ a\ \text{is}\ 2\ \text{(Right)}$\label{mcBestRuleC}
\end{enumerate}
Rule~\ref{mcBestRuleA} is responsible for pushing the car up the LHS mountain. Rule~\ref{mcBestRuleB} pushes the car right when it has a moderate amount of positive velocity and is either i) on the LHS mountain, ii) in the valley, or iii) towards the top of the RHS mountain (almost at the goal). This rule deliberately omits the case where the car is on the steeper (lower) part of the RHS mountain, because there is not enough momentum to reach the goal by pushing right. Rule~\ref{mcBestRuleC} covers this scenario, pushing the car to the right when it is on the steep part of the RHS mountain if there is a large amount of positive velocity.
\section{Conclusion}
We proposed a novel Pittsburgh GFS that utilises both MO and CoCo mechanisms to learn FRBSs that act as policies in RL environments. The system was tested on the Mountain Car environment, as it was a prime candidate due to its combination of a continuous state space and difficult exploration.
Results show that the system was able to effectively balance resources and explore the tradeoff between FRBS performance and complexity. Analysis of a selected ``best'' (near optimal performance) FRBS showed that its rules were both interpretable and parsimonious.

\bibliography{gecco2021.bib} 
\bibliographystyle{ieeetr}

\end{document}